\documentclass[10pt]{article} 

\usepackage[preprint]{rlj} 

\newcommand{\minisection}[1]{\noindent {\bf #1}}

\usepackage{amssymb}            
\usepackage{mathtools}          
\usepackage{mathrsfs}           
\usepackage{graphicx}           
\usepackage{subcaption}         
\usepackage[space]{grffile}     
\usepackage{url}                
\usepackage{lipsum}             
\usepackage[noend]{algorithmic}
\usepackage{algorithm}
\usepackage{algorithmic}
\usepackage{color} 
\usepackage{amsmath}
\usepackage{booktabs}

\title{SPEQ: Offline Stabilization Phases for Efficient Q-Learning in High Update-To-Data Ratio Reinforcement Learning}

\setrunningtitle{SPEQ: Offline Stabilization Phases for Efficient Q-Learning in High UTD Reinforcement Learning}

\author{Carlo Romeo*\textsuperscript{1}, Girolamo Macaluso*\textsuperscript{1}, Alessandro Sestini\textsuperscript{2}, Andrew D. Bagdanov\textsuperscript{1}}

\coverPageAuthor{Carlo Romeo*, Girolamo Macaluso*, Alessandro Sestini, Andrew D. Bagdanov}

\emails{\{carlo.romeo, girolamo.macaluso, andrew.bagdanov\}@unifi.it \\
asestini@ea.com}

\affiliations{
$^{1}$\textbf{MICC - University of Florence}\\
$^{2}$\textbf{SEED – Electronic Arts}\\

\par 
* The authors contributed equally as co-first authors.
}

\contribution{
We propose \textbf{SPEQ}, a novel reinforcement learning algorithm that integrates periodic offline stabilization phases to balance computational and sample efficiency.
}{
Our method contrasts with traditional high UTD ratio approaches by strategically concentrating computational resources in stabilization phases rather than uniformly distributing Q-function updates.
}

\contribution{
We empirically demonstrate that SPEQ requires from \textbf{40\%} to \textbf{99\%} fewer gradient updates and from \textbf{27\%} to \textbf{78\%} less training time compared to state-of-the-art, high UTD ratio reinforcement learning methods while maintaining competitive performance.
}{
This highlights the computational advantages of structured training schedules over conventional high UTD ratio strategies.
}

\contribution{
We show that SPEQ outperforms simple reductions in UTD ratio, demonstrating that periodic stabilization phases provide a \textbf{more effective way} to optimize learning efficiency.
}{
This distinction is crucial for designing more scalable and efficient reinforcement learning algorithms.
}

\keywords{Reinforcement Learning, UTD, computational efficiency} 

\summary{
Reinforcement learning (RL) algorithms that employ high update-to-data (UTD) ratios have demonstrated significant improvements in sample efficiency by performing multiple updates per environment interaction. However, this strategy comes at a considerable computational cost that can render it impractical for large-scale or real-world applications where efficiency is paramount. In this work we propose Offline Stabilization Phases for Efficient Q-Learning (SPEQ), a novel RL algorithm that interleaves low-UTD online training with periodic offline stabilization phases. During these phases, Q-functions are fine-tuned with very high UTD ratios while keeping the replay buffer fixed, reducing redundant gradient updates on suboptimal data. To mitigate the overestimation bias problem due to the multiple and consecutive updates, SPEQ implements dropout regularization only for critics. This approach improves computational efficiency without compromising learning effectiveness. Empirical results on the MuJoCo continuous control benchmark demonstrate that SPEQ significantly reduces computational overhead while achieving performance comparable to state-of-the-art high UTD ratio methods. 
}

\begin{document}

\maketitle  

\begin{abstract}
High update-to-data (UTD) ratio algorithms in reinforcement learning (RL) improve sample efficiency but incur high computational costs, limiting real-world scalability.
We propose Offline Stabilization Phases for Efficient Q-Learning (SPEQ), an RL algorithm that combines low-UTD online training with periodic offline stabilization phases. During these phases, Q-functions are fine-tuned with high UTD ratios on a fixed replay buffer, reducing redundant updates on suboptimal data. This structured training schedule optimally balances computational and sample efficiency, addressing the limitations of both high and low UTD ratio approaches.
We empirically demonstrate that SPEQ requires from 40\% to 99\% fewer gradient updates and 27\% to 78\% less training time compared to state-of-the-art high UTD ratio methods while maintaining or surpassing their performance on the MuJoCo continuous control benchmark. Our findings highlight the potential of periodic stabilization phases as an effective alternative to conventional training schedules, paving the way for more scalable reinforcement learning solutions in real-world applications where computational resources are constrained. 
\end{abstract}


\section{Introduction}

Reinforcement learning (RL)~\citep{sutton, survey} has gained significant attention due to its ability to solve complex decision-making tasks through interactions with environments~\citep{handManipulation, bigger}. However, one of the primary challenges in RL is sample efficiency, which is the ability to learn effectively from a limited number of environment interactions. Typically, RL requires millions of interactions with the environment to achieve strong performance, which becomes impractical in real-world applications where such interactions are expensive, time-consuming, or risky~\citep{survey}. 

Traditional off-policy RL algorithms perform a limited number of optimization updates per interaction stored in the replay buffer, leaving much of the potential learning signal unused~\citep{srsac}. Recent studies have proposed increasing the Update-To-Data (UTD) ratio -- the number of optimization steps performed per environment interaction -- as a simple yet effective strategy to address this issue~\citep{redq, droq}. By performing more updates for each experience sampled from the environment, this approach allows the agent to extract more value from each interaction, thereby improving sample efficiency.

\begin{figure}
    \centering
    \includegraphics[width=0.5\linewidth]{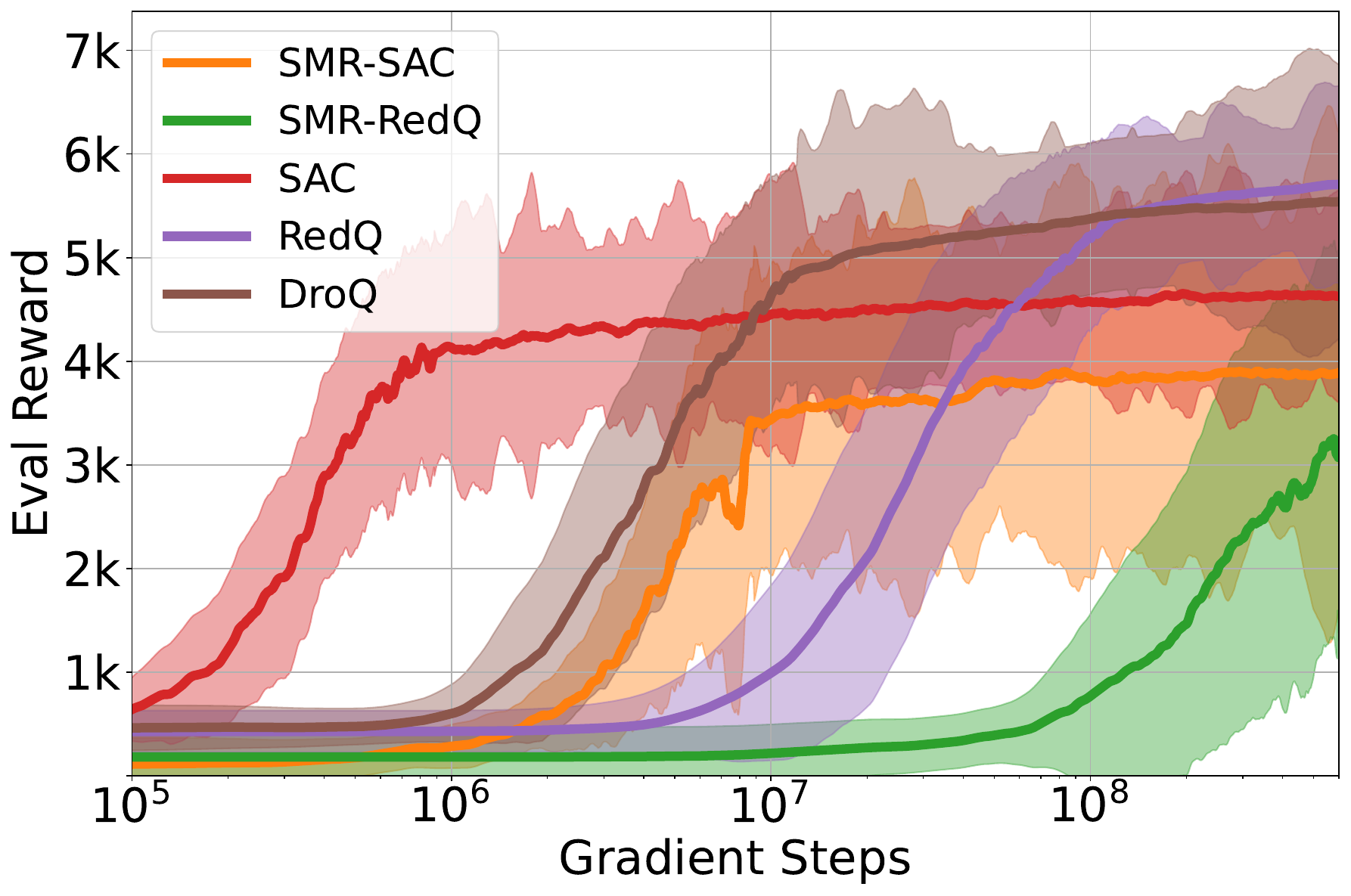}
    \caption{Comparison of state-of-the-art high UTD ratio RL approaches and SAC. This plot shows the performance averaged over four MuJoCo environments as a function of the total number of gradient steps (averaged over 5 random seeds). While high-UTD methods achieve strong final performance, they require significantly more gradient updates (and training time) compared to SAC. In contrast, SAC converges rapidly with far fewer updates, but its final performance remains limited. 
    }
    \label{teaser_plot}    
\end{figure}

However, the sample efficiency of high-UTD algorithms comes at a significant computational cost. As illustrated in Figure~\ref{teaser_plot}, these approaches typically require significantly more gradient updates, which increases the overall computational cost and training time~\citep{droq}. SAC~\citep{sac}, a low-UTD approach, converges at around one million gradient updates, achieving good performance, while high-UTD solutions such as DroQ~\citep{droq} and RedQ~\citep{redq} require one to two orders of magnitude more gradient updates, respectively, to converge to a better solution than SAC. The substantial increase in required gradient updates translates into the substantial amount of time needed for training with high UTD ratios.

This issue becomes particularly critical in scenarios involving robotic agents interacting with the real world, where computational efficiency is essential~\citep{roboticsRW, rlRWchallenges, think, expertAD, serl}. The trade-off between sample efficiency and computational cost represents a fundamental bottleneck in the scalability of reinforcement learning for real-world applications. The primary challenge lies in striking an optimal balance between fully exploiting high UTD training strategies and maintaining computational feasibility.

\begin{figure*}[t]
    \includegraphics[width=1\textwidth]{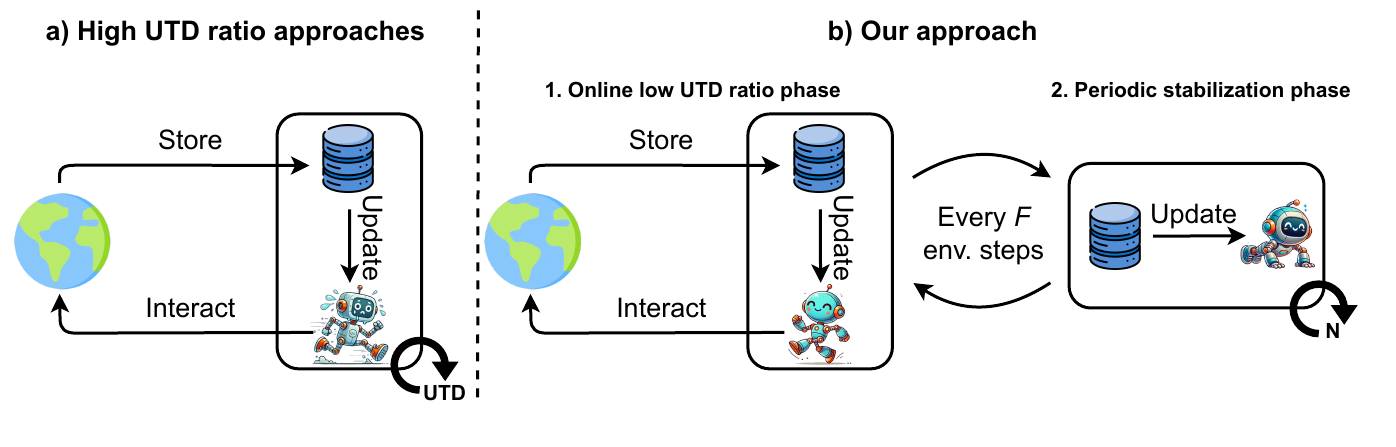}
    \caption{Overview of SPEQ. (a) Classical online RL training with high UTD ratios. For each environment interaction, the agent is trained UTD times on the replay buffer. (b) Our approach (SPEQ) which separates the training of the agent into two distinct phases. In the online interaction phase (b.1), we update the agent only once before moving to the next environment step (equivalent to $\text{UTD}=1$). Every $F$ environment steps we switch to an offline stabilization phase (b.2) in which we fine-tune the agent Q-functions for $N$ optimization steps on the current replay buffer.}
    \label{teaser}
\end{figure*}

In this paper we propose an alternative update schedule to improve the computational efficiency of off-policy RL through an increase of the \textit{performance per gradient step} value (see Figure~\ref{fig:comp}). Our approach, which we refer to as Offline \textbf{S}tabilization \textbf{P}hases for \textbf{E}fficient \textbf{Q}-Learning (SPEQ), combines the SAC~\citep{sac} algorithm with $\text{UTD}=1$ with periodic offline stabilization phases during which we interrupt online interaction with the environment, thus fixing the replay buffer, and fine-tune only the Q-functions (see Figure~\ref{teaser}). To mitigate the problem of overestimation bias, caused by the consecutive updates during offline stabilization, we incorporate dropout regularization~\citep{dropout, droq} which has been demonstrated to be more computationally efficient than large ensemble networks~\citep{redq}.

In summary, the key contributions of this work are:
\begin{itemize}
     \item We propose Offline Stabilization Phases for Efficient Q-Learning (SPEQ), a SAC variant that uses offline stabilization phases scheduled periodically during online training to achieve competitive performance while minimizing gradient updates and thus overall computational cost.
    \item We evaluate SPEQ on the MuJoCo benchmark~\citep{mujoco} and compare with state-of-the-art, high UTD ratio methods. Our experimental results show that SPEQ is significantly more efficient, performing from 40\% to 99\% fewer gradient updates and requiring from 27\% to 78\% less training time, while maintaining or surpassing the high-UTD state-of-the-art.
    \item We show that SPEQ outperforms simple reductions in UTD ratio, demonstrating that periodic stabilization phases are more effective than high-UTD reinforcement learning approaches.
\end{itemize}


\section{Related Work}
\label{sec:related}
The potential of high UTD ratios for off-policy reinforcement learning has been gaining interest from the RL research community. Model-Based Policy Optimization (MBPO) is a model-based algorithm that uses a mix of real and synthetic data along with a large UTD $\gg$ 1, achieving higher sample efficiency compared to standard model-free algorithms~\citep{mbpo}.

Randomized Ensemble Double Q-Learning (RedQ) is a model-free high UTD ratio approach using a large ensemble of Q-functions~\citep{redq}. Through careful selection of the size of the ensemble, as proposed by \citet{maxmin}, and using a random subset of the ensemble to estimate target values, \citet{redq} showed that their approach is able to minimize the expected difference between the predicted Q-values and the target Q-values (defined as the Q-function \textit{bias}). The authors showed that high-UTD algorithms reach sub-optimal performance because they are unable to cope with this bias.
RedQ is independent of the underlying optimization algorithm and can be implemented on top of any other model-free approach, such as Soft Actor-Critic (SAC)~\citep{sac}, Deep Deterministic Policy Gradient (DDPG)~\citep{ddpg}, or Twin-Delayed DDPG (TD3)~\citep{td3}. Despite its sample efficiency, the large ensemble renders the approach expensive from a computational efficiency perspective. In contrast, our method is able to alleviate the problem of increasing bias in high-UTD scenarios by using only two critics with their corresponding targets, as in classical Double Q-Learning~\citep{van2016deep} and thus avoiding a large ensemble and consequently further increasing computational efficiency.

Through the combination of dropout regularization~\citep{dropout} and layer normalization~\citep{layerNorm}, Dropout Q-Functions (DroQ) is able to leverage a smaller ensemble of Q-functions than RedQ to improve computational efficiency~\citep{droq}. Nevertheless, the total number of gradient steps required for convergence is unchanged with respect to RedQ, thus leaving room for improvement in terms of computational efficiency.

Sample Multiple Reuse (SMR) is one of the latest state-of-the-art approaches proposed to increase sample efficiency in model-free, off-policy RL~\citep{smr}. SMR applies multiple gradient steps using the same batch of transitions while avoiding overfitting thanks to the moving targets in Q-value estimation. Similarly to RedQ, SMR can be applied on top of different optimization algorithms, such as SAC and RedQ. However, the main drawback is the overall computational efficiency: in the RedQ algorithm the UTD ratio is set to 20, and, in combination with SMR, 5 more gradient steps are performed for each sampled batch. In addition, as we will see in Section~\ref{sec:experiments}, SMR is computationally less efficient than SPEQ due to the larger number of gradient updates needed.


\section{Offline Stabilization Phases for Efficient Q-Learning (SPEQ)}
\label{sec:method}

\cite{srsac} observed that, by setting the UTD ratio $\gg 1$, high-UTD approaches start to resemble Offline Reinforcement Learning~\citep{offsurvey, cql}. The problem with offline learning sessions interleaved between consecutive interactions with the environment -- a characteristic of all high-UTD approaches -- is that the replay buffer is enriched with only one new experience between one offline learning session and the next. From a computational efficiency perspective, the high-UTD approach may not be optimal due to excessive gradient updates on suboptimal data distributions. Based on these observations, the motivation behind our approach is to allow the replay buffer to grow in order to gather new and potentially more rewarding and informative experiences before investing computational resources in high-UTD learning.

We propose \textbf{SPEQ} (Offline Stabilization Phases for Efficient Q-Learning). SPEQ is a variant of SAC that interleaves low-UTD learning during online interactions with the environment with periodic offline stabilization phases. The goal of SPEQ is to optimize the computational expense of high-UTD algorithms by accentuating the offline nature of the learning process.

During the online interactions with the environment we set the $\text{UTD}=1$ to keep a one-to-one ratio between agent updates and addition of new experiences to the replay buffer and to minimize consecutive updates on similar distributions of experiences. This also helps mitigate overfitting to early-stage transitions~\citep{primacySelf, plastInjection, srsac}. After collecting enough new experiences, we switch to an offline stabilization stage that fine-tunes the critics on a fixed replay buffer.


A regularization mechanism for the critic networks is essential in SPEQ due to the very many updates performed in each stabilization phase. Conventional offline reinforcement learning methods typically employ strong regularizers (e.g. behavioral cloning~\cite{td3bc, iql, cql}) to mitigate overestimation bias. However, these approaches tend to be overly conservative in our setting and can significantly slow training~\cite{awac}, particularly since additional interactions with the environment are permitted. To address this challenge, we adopt dropout regularization~\citep{dropout}, which has been shown to effectively regulate Q-value estimates during high-UTD online training~\citep{droq}. This choice offers a computationally efficient alternative by enabling the use of only two critic networks, as in SAC, in contrast to the large ensemble architectures required in~\cite{redq} (see Section~\ref{sec:abl} in the Supplementary Materials for further analysis). Algorithm~\ref{alg1:speq} gives the pseudocode for SPEQ, with modifications from the standard SAC implementation highlighted in red.
\begin{algorithm}[t]
\caption{SPEQ}
\algsetup{linenosize=\small}
\small
\label{alg1:speq}
\begin{algorithmic}[1]
    \STATE \textcolor{red}{\textbf{Input:} Period of offline stabilization phases $\textbf{\textit{F}}$, number of  stabilization iterations phases $\textbf{\textit{N}}$} 
    \STATE Initialize policy parameters $\theta$, Q-function parameters $\phi$ and empty replay buffer $\mathcal{D}$.
    
    \FOR{ $m=1,\ldots,M$}
        \STATE Take action $a_m \sim \pi_\theta(\cdot | s_m)$. Observe reward $r_m$, next state $s_{m+1}$.
        \STATE $\mathcal{D} \leftarrow \mathcal{D} \cup (s_m, a_m, r_m, s_{m+1})$
        
        \STATE \textcolor{red}{\textbf{if} { ($m \bmod F) = 0$ } \textbf{then} $G  \leftarrow N$ \textbf{else} $G   \leftarrow 1$.}
            \FOR{ $g = 1,\ldots,G$ }
                \STATE Sample a mini-batch $\mathcal{B} = \{ (s, a, r, s') \}$ from $\mathcal{D}$.
                
                \STATE Update Q-functions $\phi$
            \ENDFOR

        \STATE Update policy $\theta$ 
    \ENDFOR
\end{algorithmic}
\end{algorithm}

\begin{figure*}
\begin{center}
    \begin{tabular}{cc} 
        \includegraphics[width=.47\linewidth]{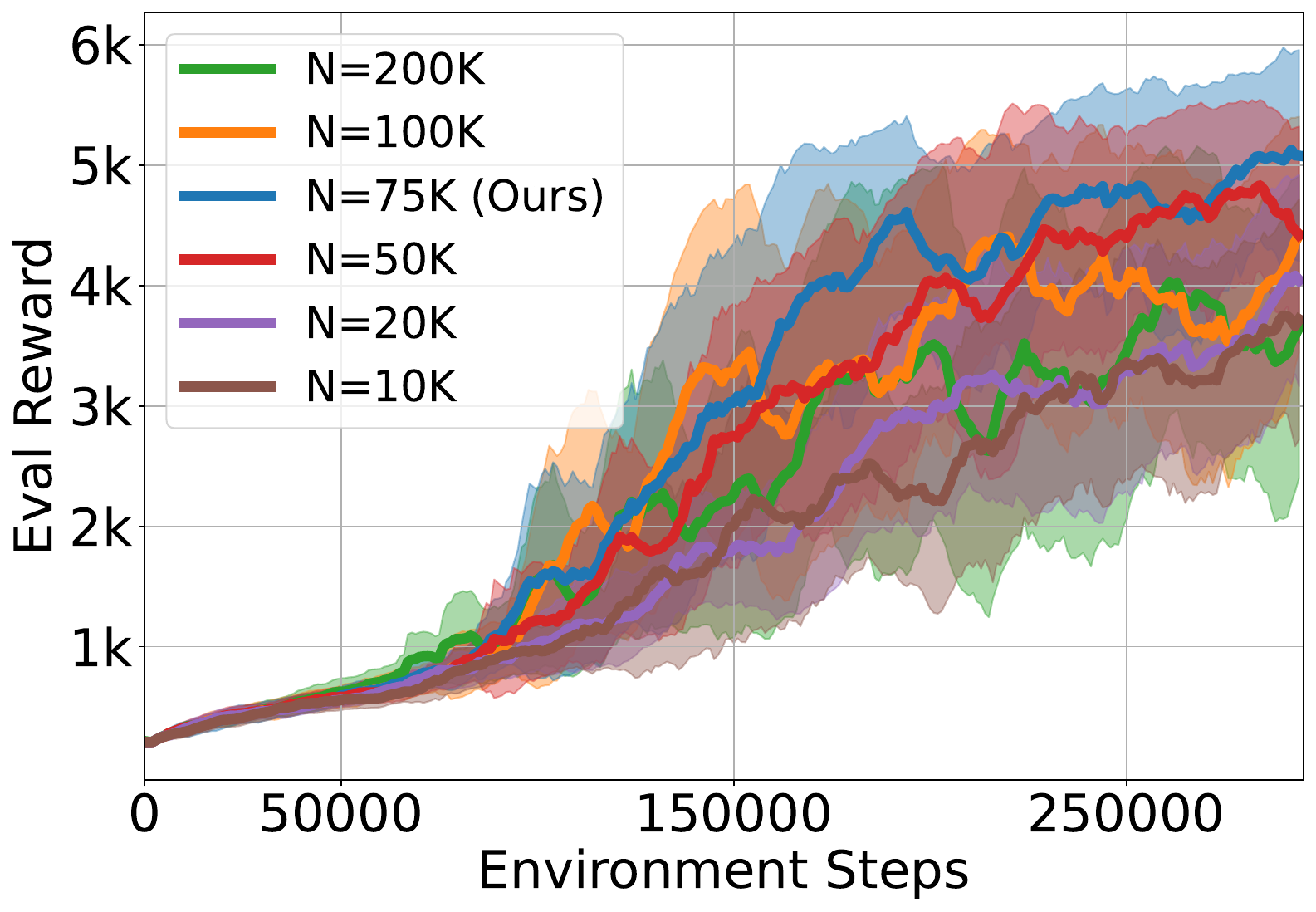}&
        \includegraphics[width=.47\linewidth]{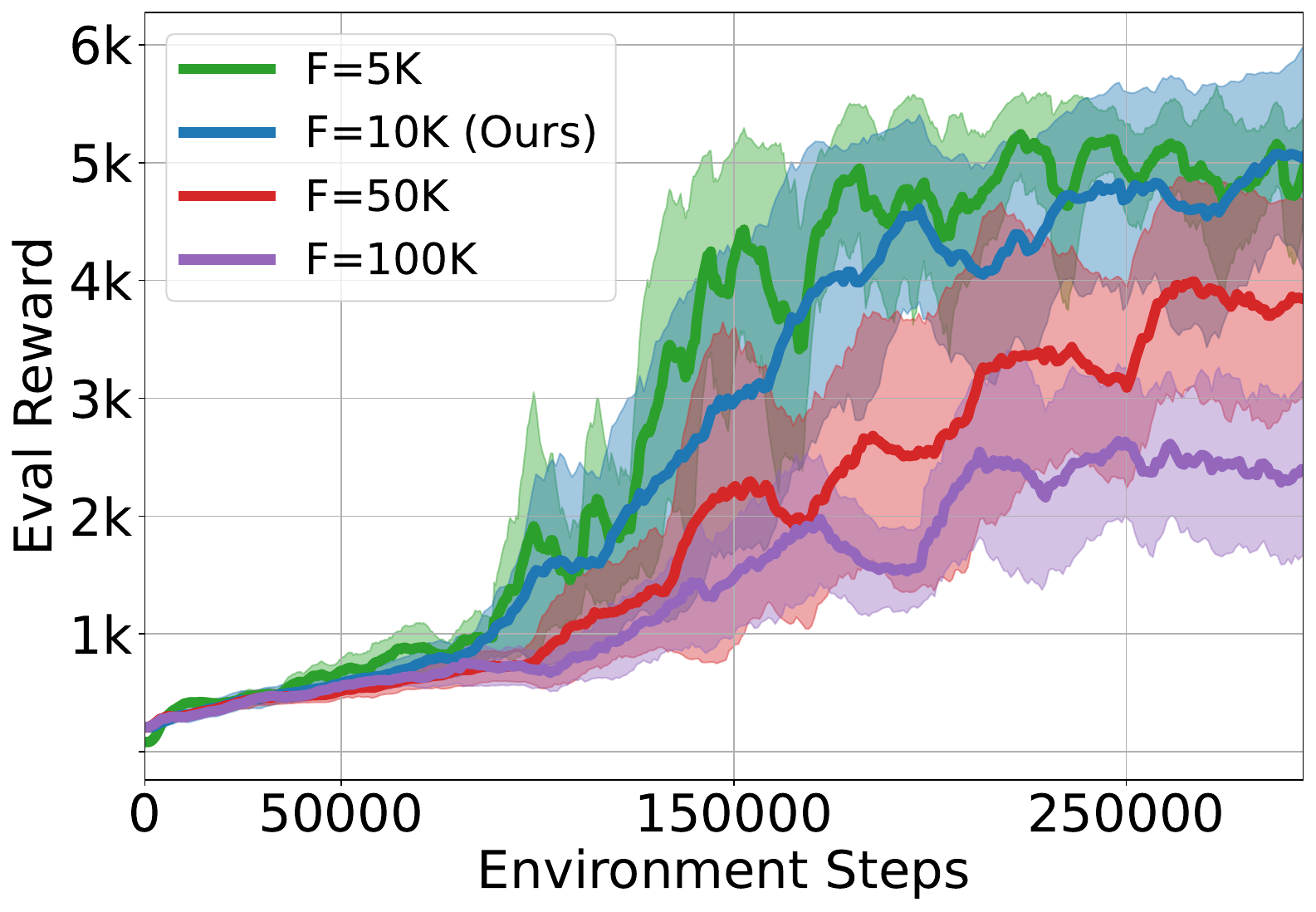}\\
         \small \hspace{0.5cm} (\textbf{a}) Length of Offline Stabilization Phases (N) \small  & \small \hspace{0.3cm} (\textbf{b})   Frequency of Offline Stabilization Phases (F)
        \end{tabular}
    \end{center}
    \caption{(a) Results of varying the number of gradient updates $N$ during offline stabilization on the MuJoCo Humanoid task, averaged over 5 random seeds. Offline stabilization phases are performed every $F=10,000$ environment steps. The plot shows that the performance improves by increasing the number of updates up to about 75K iterations, beyond which further updates result in diminishing returns. (b) Comparison of SPEQ to DroQ with varying UTD ratios. Increasing the UTD ratio in DroQ generally leads to improved performance. However, despite performing approximately the same number of gradient updates as SPEQ, DroQ with a UTD ratio of 9 results in significantly lower performance. These results indicate that reducing the UTD ratio alone significantly impacts DroQ's performance, whereas SPEQ offers a performant and computationally efficient solution.}
    
    \label{fig:combo}
\end{figure*}
\section{Experimental Results}
\label{sec:experiments}
We evaluate our approach on the OpenAI MuJoCo suite~\citep{mujoco} in following locomotion environments: Ant, Hopper, Humanoid, and Walker2d. We compare our approach with the following algorithms: SAC~\citep{sac}, REDQ~\citep{redq}, DroQ~\citep{droq} and SMR~\citep{smr}. All results are averaged over 5 random seeds. To ensure a fair comparison with published results we use the implementations of DroQ and RedQ from~\citep{droq}\footnote{https://github.com/TakuyaHiraoka/Dropout-Q-Functions-for-Doubly-Efficient-Reinforcement-Learning.git}, and the author implementation of SMR\footnote{https://github.com/dmksjfl/SMR.git}.

To verify the effectiveness of SPEQ over a span of roughly one hundred combinations of different frequencies ($F$) and number of offline steps ($N$). For brevity and clarity, we only report the most significant results. Given that SPEQ combines elements of both SAC (low UTD ratio) and DroQ (dropout regularization with high UTD updates), through our experimentation we consider SAC as a lower bound and DroQ as an upper bound in terms of computational efficiency.

Our experiments aim to answer the following research questions:
\begin{itemize}
    \item \textbf{Q1:} How do periodic offline stabilization phases affect performance?
    \item \textbf{Q2:} How frequent should offline stabilization phases be?
    \item \textbf{Q3:} How computationally efficient is SPEQ with respect to the interactions with the environment?
    \item \textbf{Q4:} How effective is each gradient step in SPEQ at increasing performance?
    \item \textbf{Q5:} To what extent does SPEQ offline stabilization improve over reducing the UTD ratio?
\end{itemize}


\begin{table}
\footnotesize
\caption{Comparison of SPEQ with state-of-the-art algorithms on 300,000 environment interactions with the MuJoCo environments in terms of total gradient steps (in millions), training time (in minutes), and final score. Compared to high UTD ratio baselines, our method significantly reduces the number of gradient steps and training time, highlighting the ability of SPEQ to balance computational efficiency against performance.}
\begin{center}
\begin{tabular}{lcccccc}
\toprule     & SAC  & SMR-SAC & DroQ & RedQ & SMR-RedQ & SPEQ (Ours) \\ \midrule
Gradient steps & 0.9 & 9.3     & 12.3 & 120  & 600  & 5.4     \\
Time             &   91     &   640   & 963  &   2100   &  1460  & 462\\ 
Score & 2894 $\pm$ 1117 & 3422 $\pm$ 1534 & 4673 $\pm$ 982 & 4923 $\pm$ 806 & 3111 $\pm$ 1989 & 4730 $\pm$871\\ \bottomrule

\end{tabular}
\end{center}
\label{tab:time}
\end{table}

\minisection{How do periodic offline stabilization phases affect performance?} To address \textbf{Q1}, we evaluate the effect of periodic offline stabilization phases in SPEQ by fine-tuning the Q-functions on a fixed replay buffer. We vary the number of gradient updates per stabilization phase while keeping the period fixed at 
twice the initial exploration phase length.
Experiments were conducted on the Humanoid task, averaging results over five seeds. To align with the computational budget of high-UTD methods, we set an upper bound of $N=200,000$ updates. As shown in Figure~\ref{fig:combo}(a), increasing the number of steps taken during the offline stabilization initially improves performance, but reaches a plateau after $N=75,000$. Further updates degrade performance due to Q-function overfitting, reducing generalization \citep{srsac}, and introducing instability rather than improving policy robustness. 

        

\minisection{How frequent should offline stabilization phases be?} To address \textbf{Q2}, we evaluate the impact of varying the stabilization phase period ($F$) while keeping the stabilization phase length constant ($N=75,000$), as identified in the previous experiment. As shown in Figure~\ref{fig:combo}(b), reducing $F$ to 5,000 maintains performance but doubles the computational cost, while increasing $F$ to 50,000 or 100,000 leads to performance degradation because of the significant reduction in the final number of stabilization steps performed. These results indicate that $F=10,000$ and $N=75,000$ provide the best trade-off between computational efficiency and learning effectiveness.
\begin{figure*}[t]
\centering
\begin{tabular}{c}
     \includegraphics[width=0.8\linewidth]{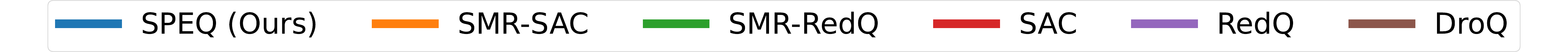}
\end{tabular}
\begin{tabular}{cc}
    \begin{minipage}{0.45\linewidth}
        \centering
        \hspace{1em} \textbf{Ant} \\
        \includegraphics[width=\linewidth]{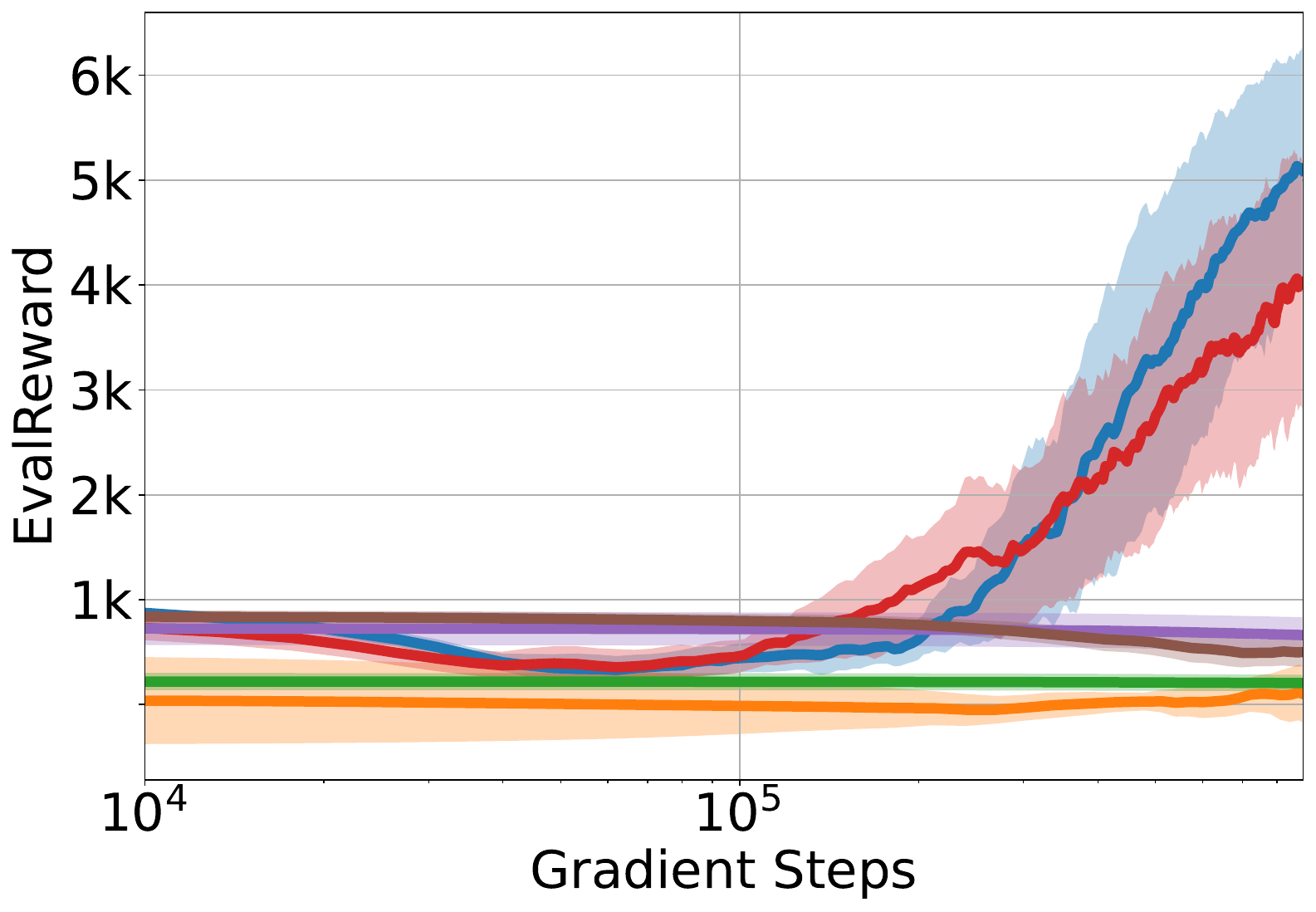}
    \end{minipage}
    &
    \begin{minipage}{0.45\linewidth}
        \centering
        \hspace{1em} \textbf{Hopper} \\
        \includegraphics[width=\linewidth]{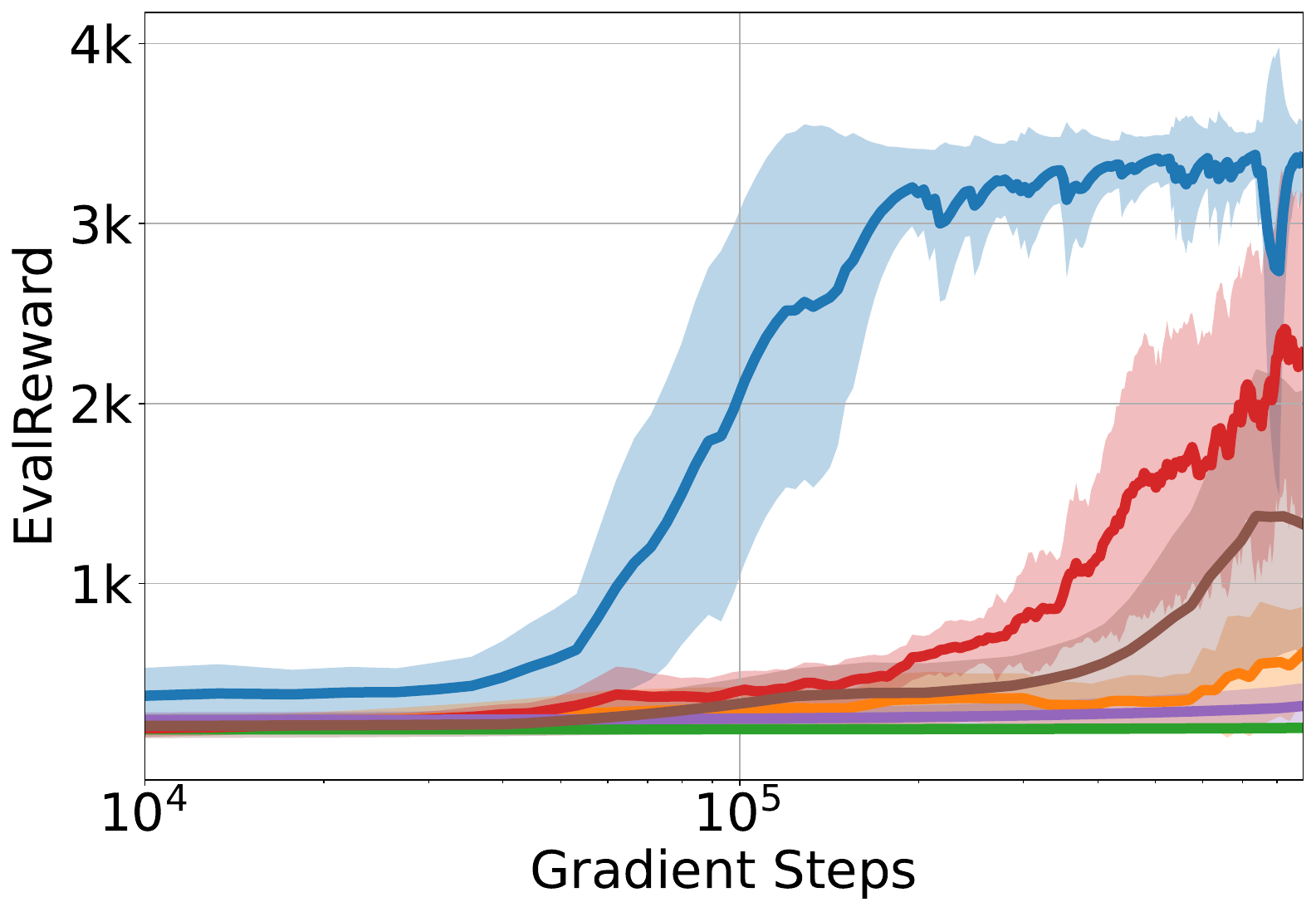}  
    \end{minipage}
    \\
    \begin{minipage}{0.45\linewidth}
        \centering
        \hspace{1em} \textbf{Humanoid} \\
        \includegraphics[width=\linewidth]{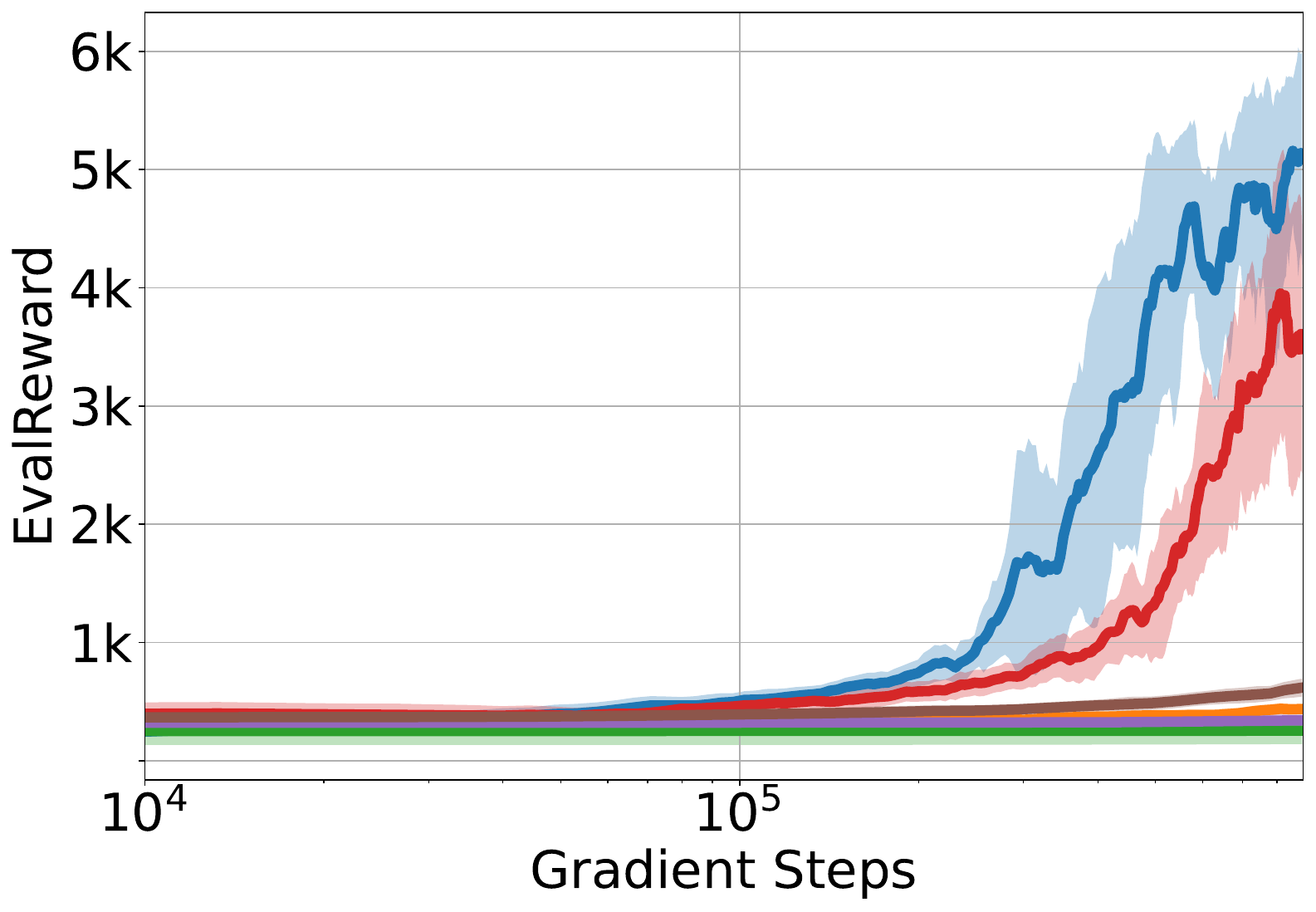}     
    \end{minipage}
    &
    \begin{minipage}{0.45\linewidth}
        \centering
        \hspace{1em} \textbf{Walker2d} \\
        \includegraphics[width=\linewidth]{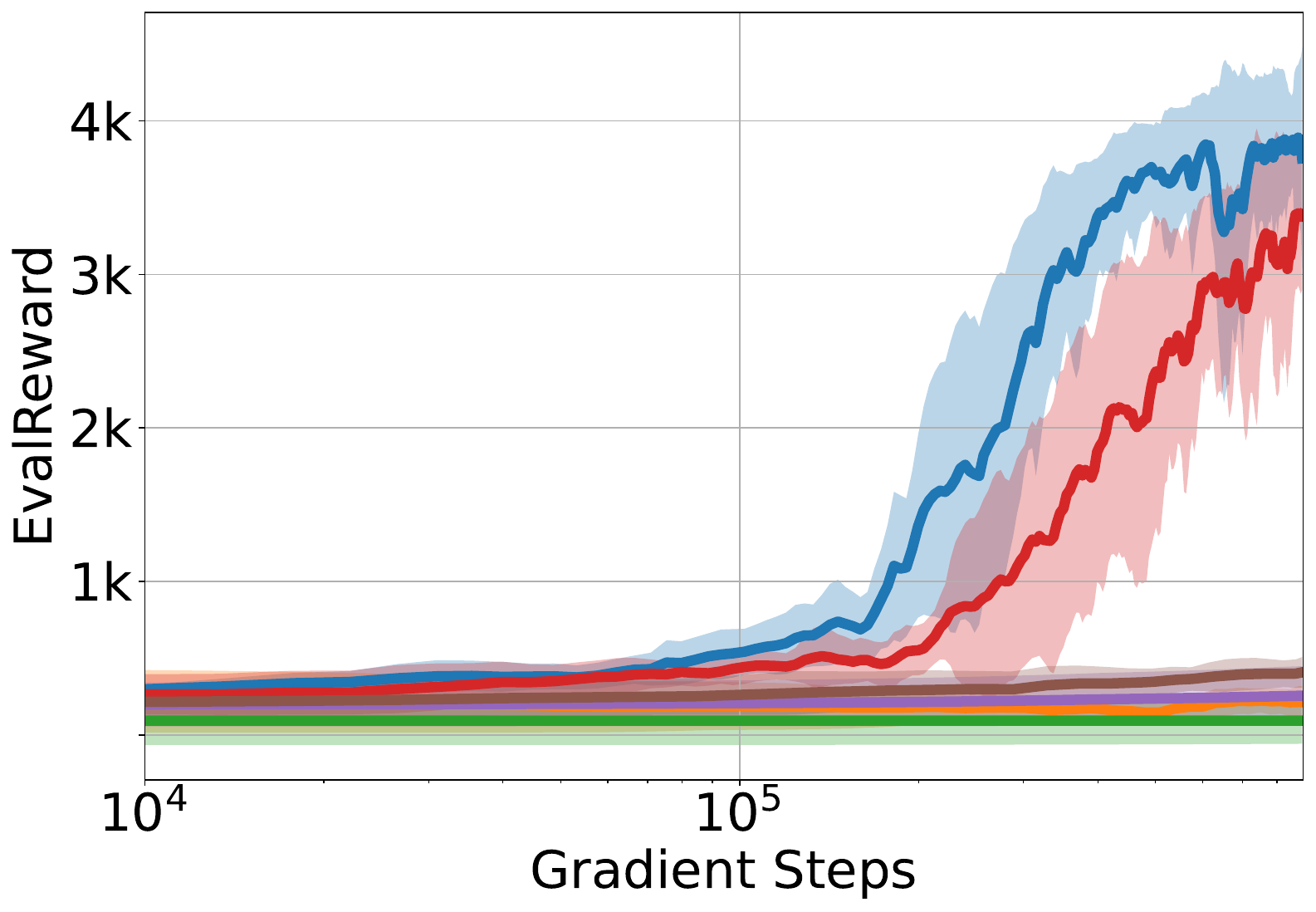}
    \end{minipage}
\end{tabular}

\caption{Comparison of SPEQ against baseline and high-UTD methods. 
Each algorithm performs the same number of gradient updates as SPEQ to evaluate the performance per gradient step value, that is: how effective each gradient step is in increasing performance in a resource-constrained scenario. We observe that high-UTD methods fail to achieve competitive performance when constrained to a limited number of updates. While SAC performs better than the high-UTD approaches, it requires more environment interactions. On the other hand, SPEQ represent the best trade-off. }
\label{fig:comp}
\end{figure*}

\minisection{Computational Efficiency.}
\label{subsec:comp}
To address \textbf{Q3}, we evaluate the computational efficiency of SPEQ compared to baseline methods. We analyze the following key factors:
\begin{itemize}
    \item \textbf{Gradient steps:} Total number of gradient updates (in millions), accounting for the number of critics used in each method. This metric is environment invariant and depends only on the algorithm used.
    \item \textbf{Time:} Average runtime per seed (in minutes) across all environments.
    \item \textbf{Score:} Final evaluation reward averaged over all seeds and environments.
\end{itemize}

Table~\ref{tab:time} gives a comparative analysis of these metrics. All experiments were performed on an Intel i7-7800X CPU and an NVIDIA GeForce GTX 1080. While SPEQ's offline stabilization phases increase total gradient steps compared to SAC, it remains significantly more computationally efficient than high-UTD approaches. Notably, SPEQ requires less than half the gradient updates of DroQ -- its closest methodological baseline -- effectively cutting training time in half. SMR-SAC falls between SPEQ and DroQ in computational cost, but achieves less than half their final performance, demonstrating an unfavorable trade-off between efficiency and performance. RedQ and SMR-RedQ are the most computationally expensive, requiring approximately 60x and 12x more gradient updates than SPEQ, respectively. 

Efficiency alone is clearly insufficient without strong performance. As shown in Table~\ref{tab:time}, SPEQ is the second best-performing method after RedQ, despite using far fewer updates, 95.5\% fewer. DroQ achieves a similar final score, but SAC and SMR-RedQ achieve about half the performance of SPEQ. While SMR-SAC may appear computationally efficient, its poor performance highlights the importance of balancing efficiency with learning effectiveness. SPEQ achieves this balance, requiring from 40\% to 99\% fewer gradient updates and from 27\% to 78\% less training time while maintaining competitive performance.


\minisection{Learning effectiveness} To address \textbf{Q4}, we evaluate the \textit{performance per gradient step value} to assess how efficiently each algorithm utilizes gradient updates when constrained to the same total number of updates as SPEQ. Unlike previous experiments where all methods were trained for 300,000 environment steps, we now standardize the training process by allowing all methods to perform exactly the same number of gradient updates as SPEQ. As shown in Figure~\ref{fig:comp}, conventional high-UTD methods like DroQ, SMR-SAC, SMR-RedQ, and RedQ truggle under this constraint, with performance remaining close to zero. This highlights the inefficiency of high UTD ratios when updates are limited, as these approaches fail to leverage their computational advantage effectively within a restricted gradient budget.

In contrast, SAC, which maintains a more balanced allocation of updates relative to replay buffer expansion, achieves a higher performance per gradient step value than high-UTD approaches, demonstrating better computational efficiency. However, SAC remains less performance than SPEQ, achieving a significant lower final score. These results emphasize the importance of strategically allocating gradient updates based on the agent’s accumulated experience.

\begin{figure*}[t]

    \begin{center}
    \begin{tabular}{c} 
        \includegraphics[width=0.6\linewidth]{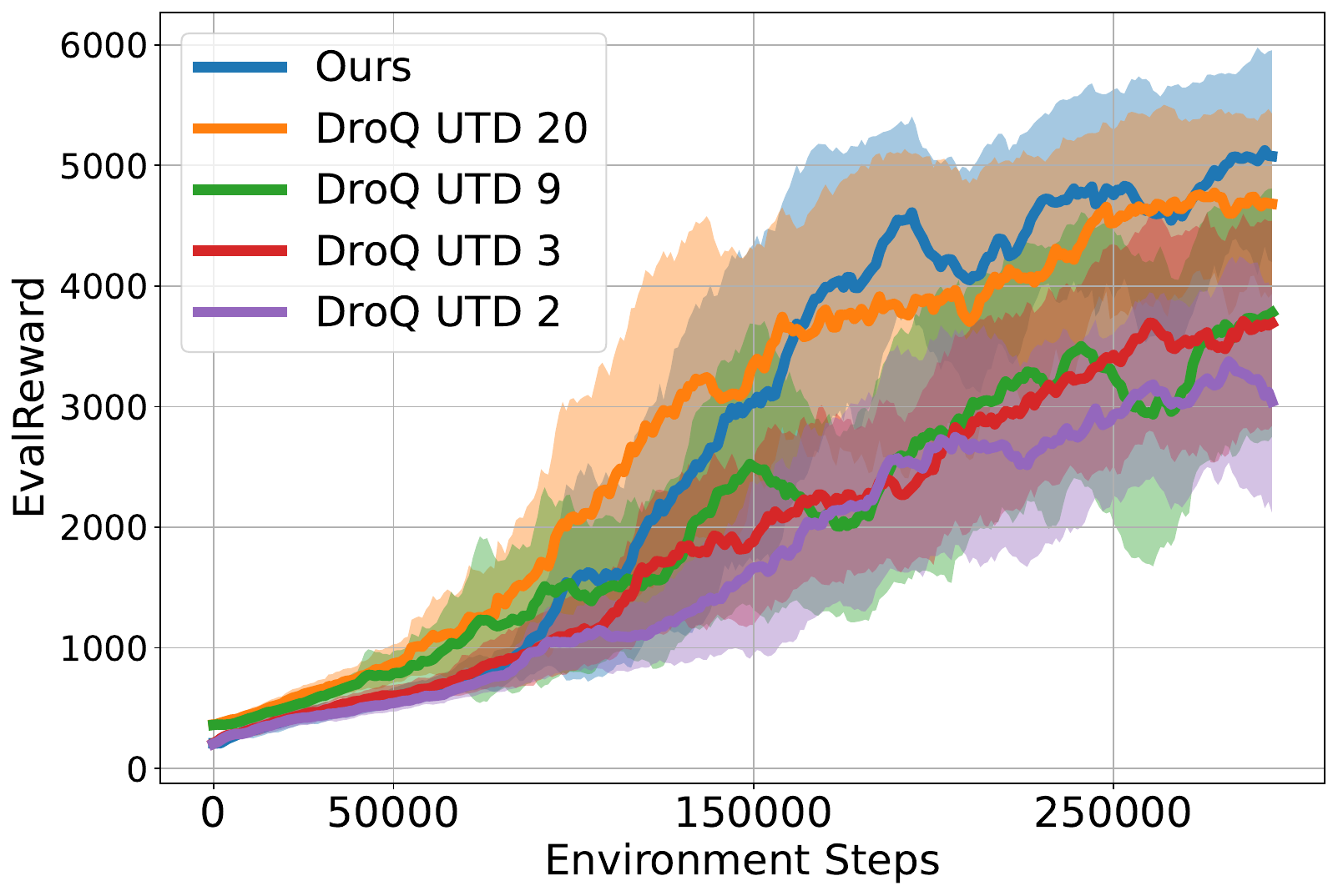}
        \end{tabular}
    \end{center}
    \caption{
    Comparison of SPEQ with DroQ at varying UTD ratios. We see that increasing the UTD ratio in DroQ generally leads to improved performance. However, despite performing approximately the same number of gradient updates as SPEQ, DroQ with $\text{UTD}=9$ results in significantly lower performance. This indicates that reducing the UTD ratio alone significantly impacts DroQ's performance, whereas SPEQ offers a more performant and computationally efficient solution.
    }
    \label{fig:abl_utd_hum}
    
\end{figure*}

\minisection{Comparison with Different UTD Ratios.} To address \textbf{Q5}, we examine whether SPEQ’s computational efficiency can be replicated by simply lowering the UTD ratio. We compare SPEQ with DroQ with UTD values of 2, 3, 9, and 20, the latter being the original DroQ setting~\citep{droq}. While DroQ with $\text{UTD}=9$ performs approximately the same number of gradient updates as SPEQ, its performance is significantly inferior, indicating that dropout regularization alone is insufficient to achieve the same tradeoff between efficiency and performance. As shown in Figure~\ref{fig:abl_utd_hum}, increasing the UTD ratio generally improves performance, with DroQ at $\text{UTD}=2$ or $\text{UTD}=3$ performing the worst. Although DroQ with $\text{UTD}=20$ achieves results comparable to SPEQ, it does so \textit{at twice the computational cost}. 
These findings highlight that SPEQ’s efficiency is not merely a result of reducing the UTD ratio, but rather stems from a structured training schedule that strategically allocates computational resources through offline stabilization phases. This demonstrates that our approach is a distinct and more effective solution for optimizing training efficiency.

\section{Conclusions}
\label{sec:conclusions}

In this work we introduced SPEQ (Offline \textbf{S}tabilization \textbf{P}hases for \textbf{E}fficient \textbf{Q}-Learning), a novel offline RL algorithm designed to improve computational efficiency while maintaining high sample efficiency. Our approach addresses the inefficiencies of conventional high UTD ratio methods by strategically interleaving low UTD online interactions with periodic offline stabilization phases. During these stabilization phases, we fine-tune Q-functions with a high UTD ratio without additional environment interactions, effectively decoupling the trade-off between computational and sample efficiency.

Through extensive empirical evaluations on the MuJoCo continuous control benchmark, we demonstrated that SPEQ significantly reduces computational costs while achieving competitive performance compared to state-of-the-art high UTD ratio methods. Specifically, SPEQ achieves 40\% to 99\% fewer gradient updates and reduces training time by 27\% to 78\%, all while maintaining or surpassing the sample efficiency of alternative approaches. Our results further highlight that offline stabilization phases are an effective alternative to simply lowering the UTD ratio, providing a structured and efficient way to allocate computational resources.
The relevance of SPEQ lies in its potential to enhance the scalability and practicality of RL. As RL is increasingly applied to real-world tasks that demand computational efficiency, methods like SPEQ provide a viable solution to mitigate excessive training costs while preserving learning effectiveness. Furthermore, our approach offers a new perspective on the design of training schedules, paving the way for more adaptive and efficient RL frameworks. 

Future work will focus on developing an automatic stabilization detection mechanism, which monitors relevant training signals, such as the evolution of the Q-value bias, policy improvement rate, or TD-error stability, to decide when and for how long offline stabilization should occur. This approach would eliminate the need for fixed hyperparameters and improve generalization across different tasks and environments.
By bridging the gap between computational and sample efficiency, SPEQ offers a promising direction for the application of reinforcement learning agents into real-world problems, such as robotics \citep{roboticsRW, serl, rlRWchallenges} and autonomous driving \citep{think, expertAD}, in which effient solutions are required.

\bibliography{main}
\bibliographystyle{rlj}



\beginSupplementaryMaterials
\section{Ablation Study}
\label{sec:abl}

This section analyzes the impact of different design choices in SPEQ. Specifically, we investigate the following key questions:
\begin{itemize}
    \item \textbf{A1:} Is regularization of the Q-functions necessary to mitigate overestimation bias during offline stabilization phases? Furthermore, is dropout regularization the most effective solution?
    \item \textbf{A2:} How does updating both the policy and Q-functions during offline stabilization phases impact performance?
\end{itemize}

\minisection{Regularization of the Q-Functions} To address \textbf{A1}, we evaluate whether dropout regularization is required to maintain stability and prevent overestimation bias in the Q-functions during offline stabilization phases. Additionally, we compare dropout against an alternative regularization strategy that employs a large ensemble of critics, as proposed in RedQ~\citep{redq}.
We consider the following variants of SPEQ for comparison:
\begin{itemize}
    \item \textbf{SPEQ w/o dropout}: SPEQ without any form of regularization.
    \item \textbf{SPEQ w/ ensemble}: SPEQ regularized using a large ensemble of critics.
    \item \textbf{SPEQ (ours)}: The proposed SPEQ variant with dropout regularization.
\end{itemize}

For completeness, we also include vanilla implementations of RedQ ($\text{UTD} = 20$) and SAC ($\text{UTD} = 1$) to provide a baseline and assess the effects of offline stabilization phases.

\begin{figure}[t]
    \centering
    \includegraphics[width=0.6\linewidth]{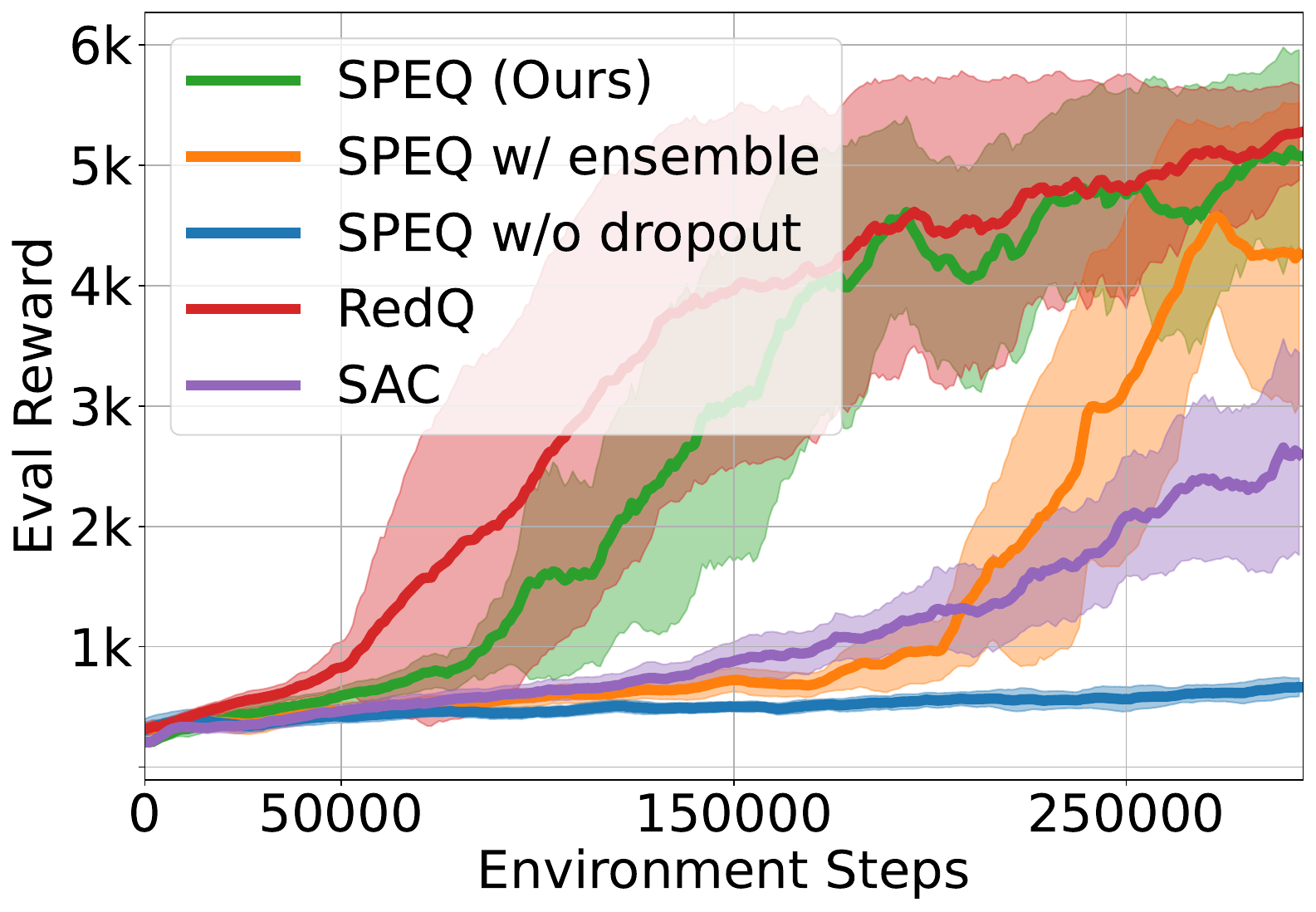}
    \caption{Comparison of different regularization techniques for mitigating overestimation bias during offline stabilization phases on the MuJoCo Humanoid task, averaged over five random seeds. 
    \textit{SPEQ (OURS)} employs dropout regularization for Q-functions, while \textit{SPEQ w/ ensemble} utilizes a large critic ensemble. \textit{SPEQ w/o dropout} does not include any regularization. 
    All SPEQ variants use the following hyperparameters: $\text{UTD} = 1, F= 10,000, N=75,000$. 
    The plot also includes RedQ ($\text{UTD} = 20$) and SAC ($\text{UTD} = 1$) as baselines, where $F$ and $N$ are set to zero.
    The results demonstrate that (i) Q-function regularization is necessary when performing multiple consecutive updates and (ii) dropout regularization outperforms ensemble-based regularization while being significantly more computationally efficient.}
    \label{fig:speq_red_sac}
\end{figure}

As shown in Figure~\ref{fig:speq_red_sac}, the absence of Q-function regularization (\textit{SPEQ w/o dropout}) leads to performance degradation, performing even worse than SAC without offline stabilization phases. This result confirms that regularization is essential when performing multiple consecutive updates on a fixed replay buffer.
Comparing dropout to ensemble-based regularization (\textit{SPEQ w/ ensemble}), we observe that dropout achieves superior performance while significantly reducing computational overhead. The ensemble approach, while effective in mitigating bias, requires a much higher number of gradient updates due to the presence of multiple critics, making it computationally expensive. Specifically, the computational cost of \textit{SPEQ w/ ensemble} scales proportionally to the number of critics (20 in the original RedQ implementation), further highlighting the efficiency of dropout regularization.

\minisection{Impact of Policy Updates During Stabilization Phases} 
To address \textbf{A2}, we evaluate different update strategies during offline stabilization phases by considering the following variants:
\begin{itemize}
    \item \textbf{SPEQ (ours)}: Only the Q-functions are updated during stabilization phases.
    \item \textbf{SPEQ w/ policy update}: Both the policy and Q-functions are updated.
    \item \textbf{SPEQ w/ only policy update}: Only the policy is updated.
\end{itemize}

\begin{figure}[b]
    \begin{center}
        \includegraphics[width=0.6\linewidth]{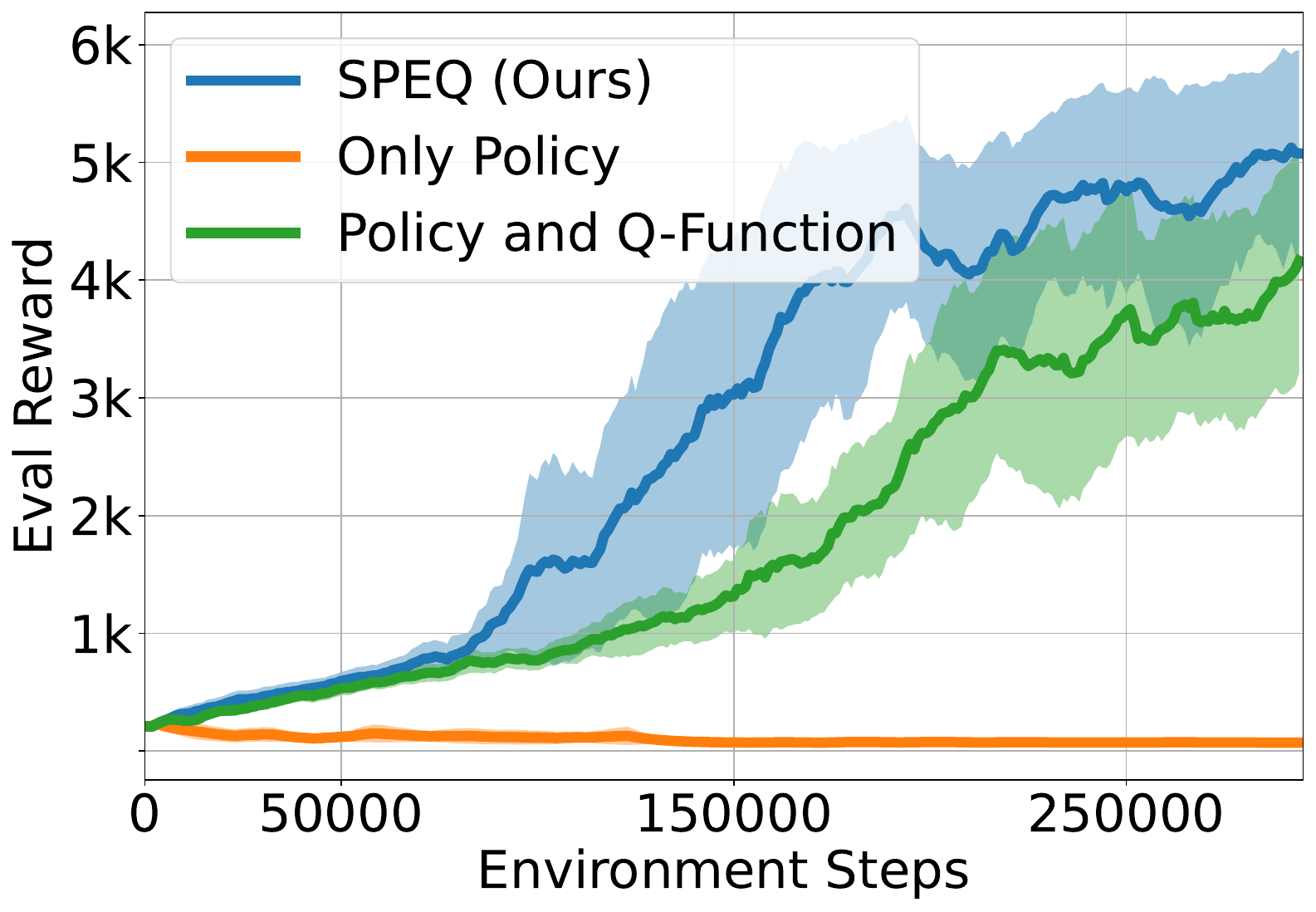}
    \end{center}
    \caption{Evaluation of different update strategies during offline stabilization phases on the MuJoCo Humanoid task, averaged over five random seeds. The blue line represents SPEQ (ours), where only the Q-functions are updated. The orange line corresponds to updating only the policy, while the green line represents updating both the policy and Q-functions.
    The results indicate that updating only the policy leads to performance collapse, while updating only the Q-functions yields the best performance and computational efficiency.}
    \label{fig:abl_pi_hum}
\end{figure}

The results, presented in Figure~\ref{fig:abl_pi_hum}, reveal two key insights: (i) Updating only the policy during offline stabilization phases leads to a \textbf{collapse in performance}, confirming that Q-function updates are crucial for effective learning. (ii) Updating both the policy and Q-functions does not yield additional benefits compared to updating only the Q-functions. Instead, it introduces additional computational overhead, making the approach less efficient.

These findings indicate that the most effective strategy is to exclusively update the Q-functions during stabilization phases, as it optimally balances performance and computational efficiency.

\end{document}